# Learning material synthesis-process-structure-property relationship by data fusion: Bayesian Co-regionalization N-Dimensional Piecewise Function Learning

A. Gilad Kusne (ORCID: 0000-0001-8904-2087), Austin McDannald (ORCID: 0000-0002-3767-926X), Brian DeCost (ORCID: 0000-0002-3459-5888)

*Abstract*—**Autonomous materials research labs require the ability to combine and learn from diverse data streams. This is especially true for learning material synthesis-process-structure-property relationships, key to accelerating materials optimization and discovery as well as accelerating mechanistic understanding. We present the Synthesis-process-structure-property relAtionship coreGionalized lEarner (SAGE) algorithm. A fully Bayesian algorithm that uses multimodal coregionalization and probability to merge knowledge across data sources to learn synthesis-process-structure-property relationships. SAGE outputs a probabilistic posterior including the most likely relationship given the data along with proper uncertainty quantification. Beyond autonomous systems, SAGE will allow materials researchers to unify knowledge across their lab toward making better experiment design decisions.**

## I. INTRODUCTION

L**ack** of advanced materials stymies many next-generation technologies such as quantum computing, carbon capture, and low-cost medical imaging. However, fundamental challenges stand in the way of discovering novel and optimized materials including 1) the challenge of a high-dimensional, complex materials search space and 2) the challenge of integrating knowledge across instruments and labs, i.e., data fusion. The first challenge arises from the need to explore ever-more complex materials as simpler material systems are exhausted. Here material system refers to the materials resulting from a set of material synthesis and processing conditions. With each new material synthesis or processing condition, the number of potential experiments grows exponentially – rapidly escape the feasibility of Edisonian-type studies, forming a high-dimensional search space. As a result, any data is typically sparse relative to the search space. The search space is also highly complex due to the underlying complex relationship between material synthesis and process conditions and the resulting material structure and functional properties, *i.e.*, the material synthesis-process-structure-property relationship (SPSPR).

Knowledge of this SPSPR plays a fundamental role across materials research, whether the research is performed by hand or through an automated or autonomous system. Researchers use knowledge of the SPSPR as a blueprint to navigate the high-dimensional complex search space toward novel and optimized materials and to explore the underlying mechanistic origins of material properties. As a result, an algorithm that properly unifies diverse materials data into SPSPR models may accelerate all these activities, impacting much of materials research. For example, such an algorithm can exploit the SPSPR to dramatically improve prediction accuracy of a target functional property, despite sparsity of data. This improved prediction would then better guide subsequent research, which would in turn boost SPSPR knowledge.

Building the SPSPR blueprint involves combining knowledge of material synthesis and process conditions, lattice structure (and potentially microstructure), as well as the diverse set of functional properties required to meet the technological requirements. This requires integrating data across different instruments and measurement modalities, each dependent on differing physical principles. Additionally, measurements can vary based on instrument calibration, measurement parameter settings, environmental conditions such as temperature and humidity, and each instrument user's measurement process. Even instruments of the same make and model differ based on unique biases, uncertainties, and data artifacts.

As a very common example, researchers often start their search for improved materials with the phase map of the target material system. A phase map (or 'phase diagram' for equilibrium materials) visualizes the synthesis-structure relationship. An example phase map is shown in Figure 1a for the $(Bi,Sm)(Sc,Fe)O_3$ material system (1). Here the phase map relates material composition (the target synthesis conditions)

*(Corresponding author: A. Gilad Kusne).*

A. Gilad Kusne is with the Materials Measurement Science Division of the National Institute of Standards and Technology, Gaithersburg MD 20899 USA and with the Materials Science and Engineering Department of the University of Maryland, College Park MD 20742 (*e-mail: aaron.kusne@nist.gov*).

Austin McDannald is with the Materials Measurement Science Division of the National Institute of Standards and Technology, Gaithersburg MD 20899 USA (*e-mail: austin.mcdannald@nist.gov*).

Brian DeCost is with the Materials Measurement Science Division of the National Institute of Standards and Technology, Gaithersburg MD 20899 USA (*e-mail: brian.decost@nist.gov*).

to resulting lattice structure, described in terms of phases, *i.e.*, composition-structure prototypes. The phase map is divided into phase regions – contiguous regions of synthesis-process space (experiments of varying synthesis and process conditions) that result in materials of the same set of phases. The regions are separated by phase boundaries (dashed black lines). Material phase information is predictive of many functional properties. Materials with property extrema tend to occur either within specific phase regions (*e.g.*, magnetism and superconductivity) or along phase boundaries (*e.g.*, caloric-cooling materials). Thus, a materials researcher can use phase maps to guide their studies toward synthesis and process conditions that are expected to produce materials with more promising properties.

Figure 1b visualizes a $(Bi,Sm)(Sc,Fe)O_3$ SPSPR by combining the phase map with the functional property of coercive electric field magnitude (CEFM) (2,3). Circles indicate experimentally characterized materials and circle color indicates measured CEFM. The CEFM is highly dependent on both synthesis conditions and phase, with the highest values occurring with 'open' hysteresis loops in the rhombohedral *R3c* phase region. Additionally, the composition dependence of CEFM significantly differs between phase regions, with greater variation occurring in the *R3c* and *Pnma* phase regions than the intermediary region. In general, discontinuities in functional property values may also occur at phase boundaries. Thus, functional properties can be represented as piecewise functions of the synthesis parameters (in this case composition), with each 'piece' of the piecewise function associated with a phase region. This allows for significant changes in function behavior from region to region and/or discontinuities to occur at phase boundaries.

For this example, data for materials synthesis and structure are used to build a phase map, and that phase map is then used to guide understanding of target property data. Knowledge is one directional, from structure to functional property. With a proper SPSPR learning algorithm, these diverse data could be combined in a unified model where knowledge of the phase map would improve analysis and prediction of functional properties and vice versa. For example, significant changes in functional properties may indicate a phase boundary and thus improve analysis and prediction of materials structure. Such an algorithm would boost overall materials research prediction accuracy and subsequent research, but such an algorithm has been lacking.

To overcome the dual challenges of a complex, high-dimensional search space and data fusion in SPSPR learning, we present the Synthesis-process-structure-property relAtionship coreGionalized lEarner (SAGE) (The SAGE algorithm is available as part of the Hermes library, https://github.com/usnistgov/hermes). SAGE is a Bayesian machine learning (ML) algorithm that combines three features: 1) ML-based segmentation of the synthesis-processing space using material synthesis, process, and phase data. Segments are phase regions, and the collection of phase regions forms the synthesis-process-phase map. The synthesis-process-phase map is then used to extrapolate the synthesis-process-structure relationship to new materials, 2) piecewise regression to fit and extrapolate synthesis-process-property relationships, and 3) coregionalization. Coregionalization allows multimodal, disparate knowledge of structure (1) and property (2), both gathered across the shared domain of synthesis and processing conditions, to be combined to exploit shared trends. SAGE combines these three features to learn the most likely SPSPR given material synthesis, process, structure, and property data. Here the language of probability is used to unify knowledge across multi-modal data with assumptions represented as priors and data combined through likelihoods. Additionally, SAGE's Bayesian framework allows for full uncertainty quantification and propagation.

Much of machine learning focuses on algorithms that provide "point estimate" outputs, i.e., they provide analysis or prediction without uncertainty. Proper uncertainty quantification and propagation requires explicitly expressing uncertainties in all variables and data, and then propagating these uncertainties through all computations to provide the uncertainty in the algorithm's outputs. A set of statistical learning algorithms such as Gaussian process regression were analytically developed to explicitly and properly manage uncertainties (4). Due to the complexity (and wide-spread use) of many algorithms, computational methods are often employed to approximate output uncertainties without significantly changing the main algorithm (5). Alternatively statistical methods such Bayesian inference can be used to build probabilistic models. With Bayesian inference, uncertainties are explicitly expressed and combined with Bayes rule to output the posterior probability (a probabilistic representation of uncertainty) – the probability distribution of the model given the data (6). When analytically intractable, sampling methods such as Markov Chain Monty Carlo (MCMC) can be used to estimate these posterior probabilities (6). Implementing these techniques is made easier through probabilistic programming languages such as Pyro and Turing (7,8). SAGE employs Bayesian inference and MCMC for uncertainty quantification and propagation.

A schematic of SAGE is provided in Figure 1c. Here SAGE takes in data streams from the material synthesis and processing systems, structure characterization instruments, as well as functional property characterization instruments. Each structure data stream is first processed using a phase analysis algorithm as described below. SAGE then learns the SPSPR from the combined phase analysis data streams and the functional property data streams. SAGE's output SPSPR posterior can be broken down into posteriors over the synthesis-process-structure phase map and the functional properties. These posteriors can then be integrated into either an experiment recommendation engine or a closed-loop autonomous materials laboratory(9), which can guide subsequent experiments and measurements in structure and functional property. For example, an autonomous system could target maximizing knowledge of the SPSPR or optimizing a material for a set of target functional properties.

Each of SAGE's features has a diverse history. The first feature of ML-based phase mapping has a seen the development of an array of algorithms over the last few decades (3,10–14). These algorithms combine two tasks, 1) data analysis: analyzing structure data to identify phase abundances or phase regions and 2) extrapolation:

extrapolating phase knowledge from measured materials to unmeasured materials. Data analysis techniques (i.e., phase or phase region identification) include matrix factorization, peak detection, graphical model segmentation, constraint programming, mixed integer programming, and deep learning, among others (3,15–23). For an example of such an algorithm applied to the provided datasets, including a thorough description of these datasets, we refer the reader to (3). Extrapolation algorithms have focused primarily on the use of graph-based models or Gaussian processes (GP) (16,17,24,25). For the present work, we assume the task of structure data analysis is addressed with one of the many available algorithms. We indicate the use of one of these algorithms with the function $m(D_s)$ applied to structure dataset $D_s$, as described below. SAGE therefore begins with knowledge of phase and focuses on the task of extrapolating phase map knowledge through Bayesian coregionalized synthesis and process space segmentation.

Piecewise function regression algorithms have a much longer history. This includes the common challenge of detecting data discontinuities - also known as jumps or changepoints, which can be generalized to higher dimensions as edges (26), change-boundaries, and change-surfaces. Changepoint detection algorithms are quite diverse, using function derivatives, filter convolution, Bayesian inference, and more recently, adaptive design (27). Common methods for piecewise regression include linear piecewise algorithms and splines. We point the reader to review articles in these fields (28,29). Specifically for GPs, multiple piecewise modeling methods exist (30) including the use of the changepoint kernel (31).

The field of coregionalization developed from geospatial science to learn functions with shared trends over the same physical domain (32,33). Data for each target function is not required to be collected for the same set of points in the input domain (33). For example, if one seeks to learn $f_1: x \rightarrow y$ and $f_2: x \rightarrow s$, data $D_1 = \{(x_k, y_k)\}_{k=1}^N$ and $D_2 = \{(x_l, s_l)\}_{l=1}^M$, the set of input locations $\{x_k\}_{k=1}^N$ and $\{x_l\}_{l=1}^M$ are not required to correspond to the same locations. Alternative methods for jointly learning related functions include multi-task learning, co-kriging, including multi-task Gaussian processes (5,34) as well as constraint programming methods and Bayesian methods (33,35,36). These algorithms focus on exploiting similarities between functions over the full underlying shared domain, assume the set of output functions are similar (*e.g.*, all continuous), and assume that each experiment is characterized similarly. Recent work tackles learning heterogenous sets of functions such as a mix of continuous, categorical, and binary outputs (37). These algorithms assume a correlation between a set of latent functions that contribute to the observed output functions.

Our challenge is unique. While we seek to jointly learn the synthesis-process-structure relationship and synthesis-process-property relationships, the correlation of interest between these relationships is purely that of discontinuities, rather than correlations over the full synthesis-process domain. We assume that phase boundaries indicate potential change surfaces in functional properties, and vice versa. We wish to jointly learn these phase boundaries and utilize them to define piecewise functions for the functional properties, allowing for different property behavior in different phase regions. Prior algorithms fail for this challenge as the synthesis-process-structure relationship and those of synthesis-process-properties are not correlated over the full synthesis-process domain (this is also true for latent property representations). Additionally, SAGE utilizes coregionalization to allow different measurements to be performed at different locations in the shared synthesis and process domain. This is commonly the case when materials synthesis and processing experiments take equal or less time than the measurements or when combining data collected at different times or by different labs.

To the authors' knowledge, the only algorithm that addresses the same challenge is the closed-loop autonomous materials exploration and optimization (CAMEO) algorithm (16). CAMEO first learns phase boundaries from synthesis, process, and structure data and then utilizes this knowledge to define the change boundaries in the piecewise function used to fit and model functional property data. This two-step approach was employed in driving an x-ray diffraction-based autonomous (robot) materials research system in the study of phase-change memory material. The study resulted in the discovery of the current best-in-class phase-change memory material – the first autonomous discovery of a best-in-class solid state material (16). SAGE improves on CAMEO by allowing full Bayesian uncertainty quantification and propagation, thus providing simultaneous information sharing between the structure and property measurements. SAGE jointly solves for the SPSPR to better exploit shared trends across structure and property data and improve SPSPR knowledge. SAGE is offered as a module of CAMEO, *i.e.*, CAMEO-SAGE.

The present data science challenge is generalizable beyond learning SPSPR. One can use SAGE to address the more common issue of having successful and failed experiments across a shared experiment parameter domain. SAGE would then learn and exploit knowledge of the success-failure boundary to improve prediction of properties of either type of experiments. Additionally, SAGE addresses data fusion across instruments, measurement modalities and labs. The common approach to this data fusion challenge is to map data from different sources into the same data space, allowing comparison. For example, data fusion for x-ray diffraction (XRD) measurements from two different XRD instruments requires removing source-based data artifacts including instrument effects that are convolved into the data. To do this, that data must then be mapped from the instrument specific, source-based independent variable space ($2\theta$) to an instrument-free independent variable space ($q$), while also accounting for differences in finite resolution in $2\theta$ space, absolute intensities and counting times, beam wavelength dispersion, and background signals, amongst other considerations. In general, data mapping to an instrument (also lab, weather, etc.) invariant space requires a significant amount of meta data that is often not available.

An alternative is to independently analyze the data from each source and then combine the derived knowledge across sources. SAGE allows such limited-metadata data fusion. The idea behind coregionalization, as implemented in SAGE, is that the boundaries identified by one measurement method are also boundaries in the other measurement methods – regardless of if those measurement methods are all nominally the same technique (*e.g.*, several different XRD instruments) or different techniques (*e.g.*, an XRD instrument and electrical coercivity measurements). For example, for structure data, one performs phase mapping analysis for each data source and then SAGE coregionalization combines knowledge across sources. A similar benefit exists for functional property data by treating data from each source as a different target property, *e.g.*, coercivity_data_source_1 and coercivity_data_source_2. Additionally, SAGE may be applied to cases where only structure data or only functional property data is obtained.

The contributions of this work are:

- Extending Bayesian coregionalization algorithms to 1-dimensional and N-dimensional joint segmentation and piecewise regression.
- Associated constraint programming algorithms for coregionalized joint segmentation and piecewise regression.
- Demonstration of Bayesian algorithms for SPSPR learning.

SAGE is a physics-informed (also known as inductive-bias informed) machine learning algorithm(38). A wide array of methods exists for integrating prior physical knowledge into machine learning methods, including engineering descriptors (39,40), latent mappings (12), constrained solution spaces (41), kernels (42), among many others. For example, a physics-informed algorithm was designed for autonomous, closed-loop control over neutron scattering to accelerate characterization of temperature-dependent magnetic structure (43). The authors represent the temperature-dependent structure as a stochastic process with neutron scattering-defined measurement uncertainties as well as a mean fraction prior defined by magnetics physics. The algorithm resulted in a fivefold acceleration in measurement efficiency. However, no previous algorithms provide the contributions listed above. Such physics-informed methods provide greater performance and lend greater interpretability to the machine learning model - providing more physically meaningful solutions.

While the provided implementation of SAGE is a surrogate model, its framework allows easy modification to embed greater prior knowledge and to increase interpretability. Target functional properties are currently defined through samples of multivariate normal distributions, similar to a Gaussian process. To increase interpretability, users can replace these samples with samples of potentially descriptive parametric models (as well as a parameter that selects between the models). SAGE will then identify the most likely model and posteriors over its parameter values. In this way a user can exploit SAGE's built-in coregionalization of functional property with phase mapping (i.e., enforced SPSPR) to boost data analysis. Additionally, one can modify parameter priors. For example, setting segmentation length scales to a Gamma distribution to increase bias for small or larger phase regions.

## II. RESULTS

We demonstrate SAGE for 1D and 2D example challenges. For both 1D and 2D, we first investigate performance for 2 edge cases, each with artificial phase maps of 2 phase regions and one artificial target functional property. In the first edge case, structure data is more informative of the change boundary and in the second edge case the functional property data is more informative of the change boundary. These edge cases demonstrate SAGE's ability to exploit knowledge across both structure and functional property data to improve prediction of both. We then provide an example of SAGE's multi-data source coregionalization capabilities with a challenge of 2 structure data sources and 2 functional property data sources. This is followed by a real-world application to the $(Bi,Sm)(Sc,Fe)O3$ and $FeGaPd$ (3) material systems.

### A. 1D Examples

The 1D challenges are shown in Figure 2 with the target functional property shown as a black dashed curve and the phase map shown as a dotted red curve that switches between a value of 0 and 1 at $x = 0.7$. For the first edge case, structure data (red diamonds) is more informative of the phase boundary, compared to functional property data (black squares). The reverse is true for the second edge case. To compare functional property prediction performance, in Figure 2a and 2b we plot: 1) SAGE-1D's functional property posterior mean (solid green line) and 95 % confidence interval (shaded green area), 2) an off-the-shelf GP with the changepoint kernel (GP-CP, blue line and shaded area) which uses maximum likelihood estimate (MLE), and 3) a plot of SAGE-1D's maximum likelihood sample (MLS, magenta line and shaded area) – the MCMC sample with the maximum computed likelihood. This sample contains an explicit changepoint value and associated piecewise GP regression. For phase boundary prediction comparison, we plot: 1) SAGE-1D's phase boundary posterior distribution (green inset histogram) and 2) SAGE-1D-PM's posterior distribution (orange inset histogram).

For the first edge case, SAGE-1D MLS combines structure and functional property knowledge to outperform GP-CP in predicting both functional property and phase boundary. SAGE-1D's slanted transition at the phase boundary (Figure 2a) indicates a range of potential phase boundary locations between the two structure data points (range is also indicated by the dotted green lines). SAGE-1D and SAGE-1D-PM have similar performance in identifying the phase boundary location, providing similar posteriors (inset). SAGE-1D employs phase boundary uncertainty to better quantify its regression uncertainty as indicated by the wider confidence intervals.

For the second edge case, SAGE-1D MLS and GP-CP have similar regression performance due to the highly informative functional property data. However, SAGE-1D outperforms SAGE-1D-PM in locating the phase mapping, as it exploits functional property data to greatly narrow in on potential locations. A further comparison between SAGE-1D, GP-CP, SAGE-1D-PM, SAGE-1D-FP, and GP classification are presented in Table 1. Knowledge of the changepoint location is limited to the two nearest data points, either functional property or structure data. As a result, functional property prediction performance is measured outside the range of the two nearest data points.

### B. 2D Examples

We observe similar behavior in the 2D demonstration of SAGE-ND as shown in Figure 3. The location of structure data (red squares on phase map plots) and functional property data (red squares on functional property plots) are indicated. For phase map prediction, SAGE-ND is compared to SAGE-ND-PM, SAGE-ND-FP, and GP classification. For functional property prediction, SAGE-ND is compared to SAGE-ND-FP and off-the-shelf GP regression. Performance scores are reported in Table 1. SAGE-ND outperforms the other methods in both phase mapping and functional property prediction for both edge cases. In edge case 2, despite highly informative functional property data, SAGE-ND outperforms off-the-shelf GP regression due to its ability to properly deal with the change in property and change in hyperparameters across the phase boundary.

In Figure 4 we demonstrate the ND algorithm for the 2D case with 2 structure data sources and 2 functional property sources. Here the first structure data source provides more information for the upper part of the phase boundary and the second source provides more information for the lower part of the boundary. SAGE-ND unifies knowledge across all four data sources to obtain good prediction of both phase map and the two functional properties.

### C. Materials Example:

For the first materials challenge demonstration, SAGE-ND is applied to learn a SPSPR for a (Bi,Sm)(Sc,Fe)O3 composition spread dataset of Raman spectra structure measurements and CEFM as shown in Figure 5. As structure data is collected primarily to learn the phase map, we present the case where structure data is more informative of the phase boundaries than the functional property data. Phase mapping and CEFM predictions estimates are shown in Figures 5a1 and 5b1 and uncertainties in Figures 5a2 and 5b2 respectively. SAGE-ND outperforms the other algorithms and shows good agreement with the ground truth (Figure 1).

For the second materials challenge demonstration, SAGE-ND is applied to learn a SPSPR for a FeGaPd (3) composition spread dataset of x-ray diffraction structure measurements and remanent magnetization as shown in Figure 6. Ground truth phase mapping and remanent magnetization are shown in Figures 6a1 and 6a2, respectively. SAGE prediction estimates are shown in Figures 6b1 and 6c1 and uncertainties in Figures 6b2 and 6c2, respectively. SAGE-ND again shows good agreement with the ground truth.

For both material systems, a comparison of SAGE-ND with SAGE-ND-PM, SAGE-ND-FP, GP classification, GP regression, and CAMEO are shown in Table 1. For the (Bi,Sm)(Sc,Fe)O3 dataset, SAGE-ND outperforms the other algorithms. For the FeGaPd dataset GPC slightly outperforms SAGE-ND on the single task of phase mapping and CAMEO performs as well as SAGE-ND on functional property prediction. However, SAGE provides proper uncertainty quantification across both phase mapping and functional property prediction, compared to CAMEO.

### III. CONCLUSION

SAGE allows one to combine knowledge of material structure and material property from multiple data sources into one joint SPSPR prediction, exploiting shared trends to maximize knowledge of the phase diagram and functional properties. The Bayesian inference methodology allows for appropriate quantification of uncertainty. By providing probabilistic descriptions of data of varying quality or fidelity (whether theory-derived or experimental), these uncertainties can then be propagated through the model by sampling the data distributions along with the model parameters and/or by replacing the piecewise GPs with heteroskedastic GPs. Additionally, correlations between functional properties can also be exploited by replacing the functional property-representing independent piecewise GPs with a coregionalized multi-output GP. These points will be the focus of future work.

Model output estimates and uncertainties can be employed in active learning-driven recommendation engines or closed-loop autonomous systems, to ensure optimum selection of subsequent experiments. For example, the phase map estimate and uncertainty can guide subsequent structure measurements toward improved phase map knowledge while the paired functional property estimates and uncertainties guide materials optimization. With each experiment increasing knowledge of separate portions of the SSPR, SAGE can play a part in unifying knowledge across a research lab toward the discovery of advanced materials.

### IV. METHODS

We present coregionalization algorithms for combining multiple data sources for materials synthesis, process, structure, and property to learn the SPSPR over the shared synthesis and processing domain $x \in X$. Structure data from data source $i$ is represented by $D_{s,i} = \{x_k, z_{k,i}\}_{k=1}^{N_i}$ for material $x_k$ (data pair indexed with $k$) and its associated structure descriptor $z_{k,i}$, with $N_i$ data pairs collected from data source $i$. The full set of structure data is labeled $D_s$, where the subscript $s$ indicated structure-associated data. Similarly, property data from data source $j$ is represented by $D_{p,j} = \{x_l, y_{l,j}\}_{l=1}^{N_j}$ for material $x_l$ and its associated material property measurement $y_{l,j}$, and where subscript $p$ indicates functional-property-associated data. $D_{s,i,k}$ and $D_{p,j,k}$ are the $i$-sourced structure data for $x_k$ and the $j$-sourced functional property data for $x_l$.

For this work we assume each data source provides data for one property. The full set of functional property data is labeled $D_p$. This representation allows for duplicate measurements of the same material from different data sources. The function $m(D_s)$ maps dataset $D_s$ to a set of phase map labels. It is one of the many such algorithms described above, and as such is not part of SAGE.

## A. Constraint Programming

The constraint programming algorithm (Eqn. 1) is defined by finding the set of parameters $\boldsymbol{\theta} = \{\boldsymbol{\theta}_s, \boldsymbol{\theta}_p\}$ that minimize the objective function $Obj$. The phase map is described by the function $f_s(\boldsymbol{x}, \boldsymbol{\theta}_s)$ which maps each point $\boldsymbol{x}$ in the target synthesis-process space $\mathbf{X}$ to a set of phase labels $\mathbf{s}$, $f_s: \boldsymbol{x} \to \mathbf{s}$, where $\boldsymbol{\theta}_s$ is the associated set of parameters. The functional property is described by the piecewise function $f_p(\boldsymbol{x}, f_s, \boldsymbol{\theta}_p)$ which maps each point $\boldsymbol{x}$ to a set of functional properties $\boldsymbol{y}$, $i.e.$, $f_p: \boldsymbol{x} \to \boldsymbol{y}$. This function is dependent on the set of parameters $\boldsymbol{\theta}_p$ and its piecewise nature is dependent on $f_s$. The functions $d_s$ and $d_p$ compute the relationship – typically the loss, between the function $f_s$ and data $D_s$ or between $f_p$ and data $D_p$, respectively. For example, $d_p$ can combine a measure of goodness of fit of $f_p$ and model complexity, $e.g.$, the Bayesian information criteria (44). To quantify loss for structure data, the data $D_s$ must also be mapped to a set of phase map labels, here performed by the function $m(D_s)$ (As discussed above, this function is one of the many found in the literature). Minimizing the objective involves: 1) identifying potential values for parameters $\boldsymbol{\theta}_s$, 2) solving for $m(D_s)$ and $f_s$, 3) identifying potential values for parameters $\boldsymbol{\theta}_p$, 4) solving for $f_p$, and 5) computing the overall loss for the objective function. This iterative approach allows a target property estimate to inform the subsequent optimization of $f_s(\boldsymbol{x})$.

$$Obj = \min_{\{\theta_s, \theta_p\}} \big[ d_s\big(f_s(\boldsymbol{x}, \boldsymbol{\theta}_s), m(D_s)\big) + d_p\big(f_p(\boldsymbol{x}, f_s, \boldsymbol{\theta}_p), D_p\big) \big] \quad (1)$$

If the loss functions are additive across datasets, we have:

$$Obj = \min_{\{\theta_s, \theta_p\}} \Bigg[ \sum_i d_s\Big(f_s(\boldsymbol{x}, \boldsymbol{\theta}_s), m(D_{s,i})\Big) + \sum_j d_p\big(f_p(\boldsymbol{x}, f_s, \boldsymbol{\theta}_p), D_{p,j}\big) \Bigg], \quad (2)$$

One implementation has $f_s$ map each point to an integer label associated with a given phase region. The function $m_s$ is then required to map the structure data to potential phase region labels similar to those of (3,10–13,15). Alternatively, one may want the overall algorithm to identify phase abundances for each material $\boldsymbol{x}$. For this case, $m_s$ identifies phase abundances and maps $\boldsymbol{x}$ to phase region labels. Abundance regression can then be performed by including abundances in the list of target properties $\boldsymbol{y}_p$.

The Bayesian model presented below can be solved using such an objective function. Here, $d_s$ and $d_p$ are the negative log likelihood functions:

$$d_s = -\ln[p(f_s(\boldsymbol{x}, \boldsymbol{\theta}_s)|m(D_s)], \quad (3)$$
$$d_p = -\ln[p(f_p(\boldsymbol{x}, f_s, \boldsymbol{\theta}_p)|D_{p,j})], \quad (4)$$

$i.e.$, the negative log likelihood of data $\{D_s, D_p\}$ being observed for functions $\{f_s, f_p\}$. This gives:

$$Obj = \min_{\{\theta_s, \theta_p\}} \big[ -(L_s + L_p) \big]$$

where $L_s$ and $L_p$ are the sum log likelihoods over all structure or all functional property observations, respectively. The notation $p()$ represents a pdf, $p(a|b)$ describes the pdf of $a$ given $b$, and for the equations below, $a \sim p(b)$ indicates drawing independent and identically distributed samples from $p(b)$. Solving for $\{\theta_s, \theta_p\}$ may be done under the variational inference approximation. The results presented her focus on Markov Chain Monte Carlo (MCMC) computed posteriors. The variational inference approximation can be used to initialize MCMC and speed up calculations.

## B. Bayesian Models

We provide two Bayesian models, one for challenges where $\mathbf{X}$ is one dimensional ($i.e.$, only one synthesis or process parameter is investigated) and one where $\mathbf{X}$ is of arbitrary dimension. Rather than minimizing loss, the aim of these models is to maximize the sum log likelihood $L$ over the set of parameters and observed data ($e.g.$, minimize sum of negative log likelihood as above). Here, MCMC is used to compute a posterior for each model parameter. Additionally, one can incorporate prior physical knowledge by modifying parameter prior probability density functions (pdf). For example, if one believes there to be many small phase regions, the uniform prior for $l_s$ can be replace with a Gamma distribution. Large expected fluctuations in a functional property can be included through modifying the $s_{r,j}$ prior**.** Both the 1-dimensional and N-dimensional models output an estimate for the posterior of the SPSPR and each parameter (given the model and data) – providing both an estimate and uncertainty, compared to the constraint programming algorithm which outputs a point estimate (estimates of the uncertainty can also be obtained). The posteriors can be used in further Bayesian analysis as demonstrated below. The MCMC Bayesian inference method for evaluating the models consists of: 1) sampling function parameters, 2) using the samples to define $f_s$ and $f_p$, and then 3) compute the log likelihood $L$.

Model 1 provides the general model. One samples the function parameter priors for $\boldsymbol{\theta}_s$ and $\boldsymbol{\theta}_p = \{\boldsymbol{\theta}_{p,j,r}\}$ for each $j$ of $J$ functional properties (or function property data source) and each $r$ of $R$ phase regions. $f_{p,j}$ is a piecewise random process with different behavior $f_{p,j,r}$ for each functional property in each phase region, i.e., different kernel hyperparameters for each phase region. $f_s$ is used to compute the categorical distribution $p(r(\boldsymbol{x}))$ of phase regions labels for each point $\boldsymbol{x}$. $p(r(\boldsymbol{x}))$ is used to compute the sum log likelihood $L_s$ of structure data observations and identify phase region label probabilities for each functional property observation data point $x_p$. The sum log likelihood of the observed functional properties $L_p$ is computed using these probabilities and the piecewise $f_{p,j}$. The total likelihood $L$ is then returned, guiding Bayesian inference sampling. The implementations and associated code can be used with an arbitrary number of data sources. Sampling from GPs uses the Cholesky decomposition method to improve MCMC stability (4).

| **Model 1** General SAGE Model |
| --- |
| 1    $\boldsymbol{\theta}_s \sim prior_{\theta_s}$ |
| 2    $p_r(r) = p_r(r(\boldsymbol{x}) = l) = \text{Categorical}\big(f_s(\boldsymbol{\theta}_s)\big)$ |
| 3    $L_s = \sum_i \sum_k \ln\left[p\left(m(D_{s,i,k})|p_r(r)\right)\right]$ |
| 4    $\boldsymbol{\theta}_p = \{\boldsymbol{\theta}_{p,j,r}\}_{j=1,r=1}^{j=J,r=R} \sim prior_{\theta_{p,j,r}}$ |
| 5    $f_{p,j} = \sum_r f_{p,j,r}(\boldsymbol{\theta}_{p,j,r}) p_r(r)$ |
| 6    $L_p = \sum_j \sum_k \ln\left[p\left(D_{p,j,k}|N(f_{p,j},\boldsymbol{\theta}_p)\right)\right]$ |
| 7    $L = L_s + L_p$ |

After Bayesian inference is run, *i.e.*, each $b$ sample of $B$ total MCMC samples are collected, the Bayesian posteriors for the phase map and functional-properties-describing functions are approximated. Here the categorical distribution describing the phase map is computed by taking $\text{mean}_b[p_b(r)]$, the posterior mean over the sampled categorical distributions. The phase map estimate $\hat{p}$ and uncertainty $e_{\hat{p}}$ are then computed with $\hat{p} = \text{argmax}_r[p_{\boldsymbol{M}}]$ and $e_{\hat{p}} = \text{entropy}_r[p_{\boldsymbol{M}}]$. Each functional property is described by the posterior multivariate normal distribution $N\big(\text{mean}_b[f_{p,j,b}], \text{std}_b[f_{p,j,b}]\big)$ with additional measurement noise $\text{mean}_b[n_{p,j,b}]$.

| **Algorithm 1** Post MCMC Bayesian Analysis |
| --- |
| For the Bayesian inference sample index $b$: |
| 1    $p_{\boldsymbol{M}} = \text{mean}_b[p_b(r)]$ |
| 2    $\hat{p} = \text{argmax}_r[p_{\boldsymbol{M}}]$ |
| 3    $e_{\hat{p}} = \text{entropy}_r[p_{\boldsymbol{M}}]$ |
| 4    $\hat{\mu}_{p,j} = \text{mean}_b[f_{p,j,b}]$ |
| 5    $\hat{\sigma}_{p,j} = \text{standard\_deviation}_b[f_{p,j,b}]$ |
| 6    $\hat{n}_{p,j} = \text{mean}_b[n_{p,j,b}]$ |
| 7    $\hat{f}_{p,j} = N(\hat{\mu}_{p,j}, \hat{\sigma}_{p,j})$ |
| 8    $\hat{y}_{p,j} = N(\hat{\mu}_{p,j}, \hat{\sigma}_{p,j} + \hat{n}_{p,j})$ |

The SAGE algorithms make use of latent functions. One set of latent functions are used to identify the probabilities of each point $x$ belonging to a specific phase region. These probabilities are then multiplied by an additional set of latent functions describing target functional properties, in effect weighting these second set of functions to bound them to target phase regions. Through this combination of latent functions, one can identify regions in $X$ that may contain significant changes in phase and/or functional properties and may be of interest for further experiments. Statistical analysis of multiple samples of latent functions provides a posterior distribution for phase map and piecewise functional properties.

## A. One Dimensional Challenges

When $\mathbf{X}$ is one dimensional, phase boundaries may be represented as change points. The set of structure model parameters $\boldsymbol{\theta}_s$ are simply a set of changepoints in $\mathbf{X}$. The changepoints $\boldsymbol{\theta}_s$ are sampled and then converted to categorical distribution $p(r) = f_s(\boldsymbol{x})$. Each continuous region of $\mathbf{X}$, bounded by either a change point or the edge of the search space, defines a phase region $r$. For example, for a 2-phase region challenge over $\mathbf{X} = [0,1]$ with one changepoint at arbitrary value 0.5, phase region 0 would have a probability of 1 at $x = < 0.5$ and a probability of 0 for $x > 0.5$ and vice versa for phase region 1. The categorical distribution is then used to compute the likelihood of the observations given the samples.

The presented implementation is developed from that of (25). The functional property in each phase region is represented by an independent radial basis function kernel Gaussian process, with $\boldsymbol{\theta}_p$ including: $l_{r,j}$ kernel length scale, $s_{r,j}$ kernel standard deviation (also known as 'scale'), and $s_j$ measured noise standard deviation. For this work, we assume that $s_j$ is the same for property $j$ across all phase regions. For each property, the region-specific functions $f_{p,j,r}$ are sampled from $GP(\boldsymbol{\theta}_{p,j,r})$ and then combined using the probabilistic weights $p(r)$ to give the piecewise functions $f_{p,j}$. $f_{p,j}$ describes the sample mean and $n_j$ the sample noise of the multivariate distribution $N(f_{p,j}, n_j)$ used to describe a potential generating random process. Data likelihood is then given by $p\left(D_{p,j}\middle|N(f_{p,j}, n_j)\right)$.

Example Implementation:

| **Model 2**: 1-Dimensional SAGE |
| --- |
| 1    $\boldsymbol{\theta}_s \sim \text{Uniform}(X)$ |
| 2    $\boldsymbol{M}_s = \text{membership}(\boldsymbol{\theta}_s, \boldsymbol{x}_s)$ |
| 3    $p_r(r) = p_r(r(\boldsymbol{x}) = l) = \text{Categorical}(\boldsymbol{\theta}_s)$ |
| 4    $L_s = \sum_i \sum_k \ln\left[p\left(m(D_{s,i})|p_r(r)\right)\right]$ |
| 5    $l_{r,j} \sim \text{Uniform}(\min\_length\_scale_{r,j}, \max\_length\_scale_{r,j})$ |
| 6    $s_{r,j} \sim \text{Uniform}(\min\_standard\_deviation_{r,j}, \max\_standard\_deviation_{r,j})$ |
| 7    $n_j \sim \text{Uniform}(\min\_noise\_scale_j, \max\_noise\_scale_j)$ |
| 8    $\boldsymbol{\theta}_p = \{l_{r,j}, s_{r,j}, n_j\}$ |
| 9    $f_{p,j,r} \sim GP_{j,r}(\boldsymbol{\theta}_p)$ |
| 10   $f_{p,j} = \sum_r f_{p,j,r}\, p(r)$ |
| 11   $L_p = \sum_j^J \sum_r^r \ln\left[p\left(D_{p,j}\middle|N(f_{p,j}, n_j)\right)\right]$ |
| 12   $L = L_s + L_p$ |

**Model 3**: N-Dimensional SAGE

1  $\boldsymbol{l}_s \sim \text{ND\_Uniform}(\text{min\_length\_scale}_s, \text{max\_length\_scale}_s)$

2  $\boldsymbol{s}_s \sim \text{ND\_Uniform}(\text{min\_standard\_deviation}_s, \text{max\_standard\_deviation}_s)$

3  $\boldsymbol{\theta}_s = \{\boldsymbol{l}_s, \boldsymbol{s}_s\}$

4  $\boldsymbol{W}_k(\boldsymbol{x}) = \{w_h(\boldsymbol{x})\}_{h=1}^R \sim N\left(0, K_{Matern\ 5/2}(\boldsymbol{x}, \boldsymbol{x}', \boldsymbol{\theta}_s)\right)$

5  $p(r) = p(r(\boldsymbol{x}) = l) = \exp w_{r=l}(\boldsymbol{x}) \,/\, \sum_r \exp w_r(\boldsymbol{x})$

6  $L_s = \sum_i \sum_k \ln\left[p\left(m(D_{s,i})|p(r)\right)\right]$

7  $\boldsymbol{l}_{r,j} \sim \text{ND\_Uniform}(\text{min\_length\_scale}_{r,j}, \text{max\_length\_scale}_{r,j})$

8  $\boldsymbol{s}_{r,j} \sim \text{ND\_Uniform}(\text{min\_standard\_deviation}_{r,j}, \text{max\_standard\_deviation}_{r,j})$

9  $n_j \sim \text{Uniform}(\text{min\_noise\_scale}_j, \text{max\_noise\_scale}_j)$

10  $b_{j,r} \sim \text{Uniform}(\text{min\_bias}_{r,j}, \text{max\_bias}_{r,j})$

11  $\boldsymbol{\theta}_p = \{\boldsymbol{l}_{r,j}, \boldsymbol{s}_{r,j}, n_j\}$

12  $f_{p,j,r} \sim N\left(b_{j,r}, K_{RBF}(\boldsymbol{x}, \boldsymbol{x}', \boldsymbol{\theta}_p)\right)$

13  $f_{p,j} = \sum_r f_{p,j,r}\, p(r)$

14  $L_p = \sum_j \sum_r \ln\left[p\left(D_{p,j} \big| N(f_{p,j}, n_j)\right)\right]$

15  $L = L_s + L_p$

### B. N-Dimensional Challenges

Change boundaries and surfaces are not easy to define in higher dimensions, so we instead sample latent functions $w(\boldsymbol{x})$ and then transform these latent functions with categorical distributions for phase region labels over $\mathbf{X}$. We define an N-dimensional multivariate normal distribution for the latent functions with associated parameters $\boldsymbol{\theta}_s$. For an SPSPR with R phase regions, we take R latent function samples, again, using the Cholesky decomposition method. The samples $w(\boldsymbol{x})$ are then used to define the columns of matrix $\boldsymbol{M}_s$, i.e., $\boldsymbol{M}_s[:, r] = w_r$. Each entry of $\boldsymbol{M}_s[k, r]$ is taken as the unnormalized event log probability (and converted to logits by the Categorical distribution function) for point $\boldsymbol{x}_k$ belonging to phase region label $r$. Here, each functional property is described by a N-dimensional GP.

### C. Additional Models

In this work, we compare SAGE to a set of algorithms including off-the-shelf GP regression and classification, modified versions of SAGE, and CAMEO. As the space of machine learning algorithms is vast, we down selected comparisons for the following reasons. There are no other algorithms that perform the same joint task as SAGE, i.e., coregionalized joint segmentation and piecewise regression from disparate classification and regression datasets. However, CAMEO works toward the same multitask goal through non-joint learning, and as a result, is one benchmark algorithm. SAGE is also compared to off-the-shelf GP regression and classification algorithms as it shares the assumptions of each of these algorithms, though SAGE also contains the assumption of coregionalization across data sources. By benchmarking against these algorithms, we demonstrate SAGE's improvements over algorithms with shared set of (reduced) assumptions. Similarly, we benchmark SAGE's benefits of coregionalization against limited versions of SAGE, e.g., where SAGE is provided data from only one of the data sources.

We compare SAGE to off the shelf GP algorithms and modified versions of SAGE. We compare SAGE's phase mapping (PM) performance with a version of SAGE which only takes structure data input. For 1D challenges this is Model 4 'SAGE-1D-PM' and for 2D challenges this is Model 6 'SAGE-ND-PM'. We compare SAGE's functional property (FP) prediction performance with versions that only take in functional property data, i.e., piecewise Gaussian process regression. For 1D challenges this is Model 5 'SAGE-1D-FP' and for 2D challenges Model 7 'SAGE-ND-FP'. For these algorithms that rely on just one input data type, it is expected that for exhaustive data, performance will be high, while for partial data, the joint SAGE model will outperform these models. These additional algorithms are available as part of the SAGE library.

- Model 4, SAGE-1D-PM: This algorithm mirrors SAGE-1D but excludes functional property regression. The algorithm is described the same as Model 2 lines 1-4 and returns $L_s$.

- Model 5, SAGE-1D-FP: This algorithm mirrors SAGE-1D but excludes the phase mapping loss term. It is thus a 1-dimensional piecewise GP. The algorithm is described the same as Model 2 lines 1-3 and 5-11 and returns $L_p$.

- Model 6, SAGE-ND-PM: This algorithm mirrors SAGE-ND but excludes the task of functional property regression. The algorithm is described the same as Model 3 lines 1-6 and returns $L_s$.

- Model 7, SAGE-ND-FP: This algorithm mirrors SAGE-ND but excludes the phase mapping loss term. It is thus an N-dimensional piecewise GP. The mode is described the same as Model 3 lines 1-5 and 7-14 and returns $L_p$.

- GP-CP; GPR; GPC: The implementations use the radial basis function kernel for regression and the Matern 5/2 kernel and MultiClass likelihood for classification. All use the truncated Newton method for optimization.

- CAMEO – Only piecewise regression task. This model follows that of []. A Gaussian random field (GRF) is defined for the material system including both characterized and potentially characterized materials. The GRF is applied to the structure data to segment the material system and that segmentation is then combined with off-the-shelf Gaussian process regression, using different hyperparameters for each phase region.

### D. Implementation

The provided SAGE implementations are designed for parallel computation across systems with multiple CPUs, allowing for easy scalability for large datasets. Implementations include: SAGE-1D; SAGE-ND for one structure data stream input and multiple functional property data stream inputs, where the functional property data streams are measured over the same materials (though potentially different materials than the structure data stream); and SAGE-ND-MULTI for multiple structure and functional property data streams where materials investigated can be different for all data streams.

SAGE was run on a laptop (6 core 2.7GHz, 32GB memory, NVIDIA* Quadro P620) and runs within a few minutes, e.g., less than 2 minutes for the (Bi,Sm)(Sc,Fe)O3 material system example. All implementations are built to boost performance through parallelization across multiple CPUs by changing the "number of available cores" and "number of chains". This dramatically accelerates computation. For example, using parallelization across a 100 CPU node allows MCMC samples for each CPU to be reduced by an order of 100.

Here we provide initial values or uniform prior ranges for the implementation. If a parameter is not mentioned, it is the default initial value or range of the library used.

- GPs
  - All GP implementations are written in gpflow (45).
  - 1D CP-GP: initial length scale = .2; initial change point steepness = 100; noise variance = 0.01; max iterations =10000;
  - 2D GPR: initial lengthscales = 1.; noise variance = 0.005 and range [0.001, 0.01]; max iterations = 1000;
  - 1D and 2D GPC: max iterations = 1000;
- All MCMC algorithms
  - Number warmup samples = 100; number samples = 1000; target acceptance probability = 0.8; max tree depth = 5; jitter = 1E-6
  - SAGE-1D, SAGE-1D-PM, SAGE-1D-FP are written in numpyro (46) with parameters: $s_{r,j} = [.01, 2.]$; $l_{r,j}$ [.2,1.]; $n_j = [0.001,.01]$; Change point bounds, i.e., $\theta_s = [0.5, 1.]$
  - SAGE-ND, SAGE-ND-PM, SAGE-ND-FP are written in numpyro Jax with parameters
    - SVI initialization of phase map: number of samples = 100000; Adam step size = 0.05;
    - MCMC algorithm: $s_s = [5.,10.]$; $l_s = [.1,2.]$; $s_{r,j} = [.1, 2.]$; $l_{r,j} = [1,2.]$; $n_j = [0.001,.1]$; $b_{r,j} = [-2., 2.]$
  - SAGE-ND Multiple data sources:
    - SVI initialization of phase map: number of samples = 10000; Adam step size = 0.01; $s_s = [5.,10.]$; $l_s = [1.,2.]$;
    - MCMC algorithm: number of warmup steps: 100; number of samples: 2000; number of chains = 100; $s_s = [5.,10.]$; $l_s = [.1,2.]$; $s_{r,j} = [.1, 2.]$; $l_{r,j} = [.1,5.]$; $n_j = [0.001,.1]$; $b_{r,j} = [-2., 2.]$
- CAMEO: Uses the same parameters as in (16).

* NIST DISCLAIMER: Certain commercial equipment, instruments, or materials are identified in this paper to foster understanding. Such identification does not imply recommendation or endorsement by the National Institute of Standards and Technology, nor does it imply that the materials or equipment identified are necessarily the best available for the purpose.

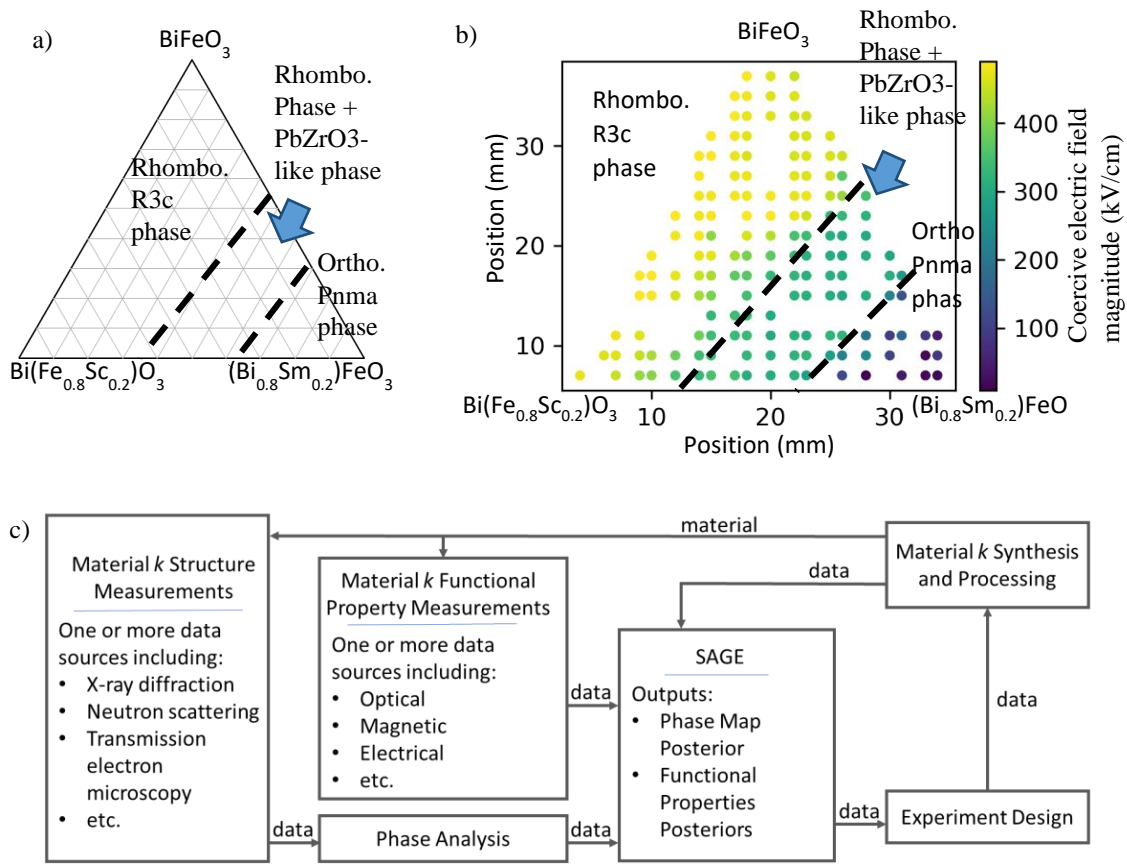

**Fig. 1**. a) The (Bi,Sm)(Sc,Fe)O₃ material system experimentally identified phase diagram. Phase boundaries indicated by black dashed lines. b) (Bi,Sm)(Sc,Fe)O₃ coercive electric field magnitude overlayed with phase diagram. Circles indicate experimentally characterized materials and color indicates coercive electric field magnitude between 0 kV/cm and 491 kV/cm. c) SAGE Schematic. A collection of materials spanning a target material system are characterized for multimodal structure data which is then processed through a preliminary phase analysis tool. The materials are also characterized for a range of functional properties. The collected data is passed to SAGE which outputs posterior probabilities for both the material system phase map and the functional properties. These posteriors can then be used in an experiment design (active learning) algorithm to determine the next material to investigate.

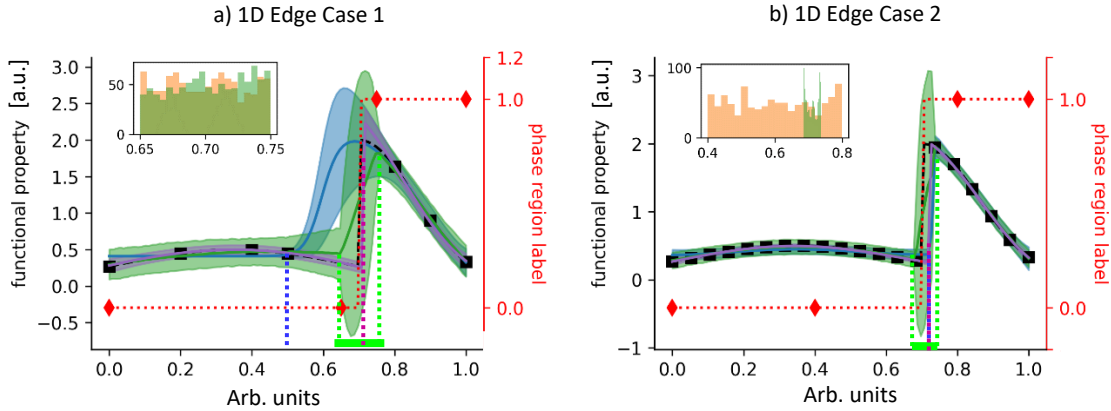

**Fig 2.** SAGE-1D performance for two edge cases: a) edge case 1 where structure data (red diamonds) is more informative of the phase boundary and b) edge case 2 where functional property data (black squares) is more informative of the phase boundary. In both parts, the main plots show comparison of: the SAGE-1D posterior mean and 95 % confidence interval (green line and shaded region), and SAGE-1D maximum likelihood sample (magenta line and shaded region), and a GP (blue line and shaded region) with a changepoint kernel and two radial basis function kernels on either side of the changepoint. The ground truth phase map is indicated by a red dotted line and the ground truth functional property function is indicated as a black dashed line. The GP identified changepoint is indicated with a blue dotted line and the range of potential changepoints identified by SAGE-1D is indicated by green dotted lines. The SAGE-1D changepoint posterior is shown as the inset (green histogram) and compared to the SAGE-1D-PM changepoint detection algorithm (orange histogram).

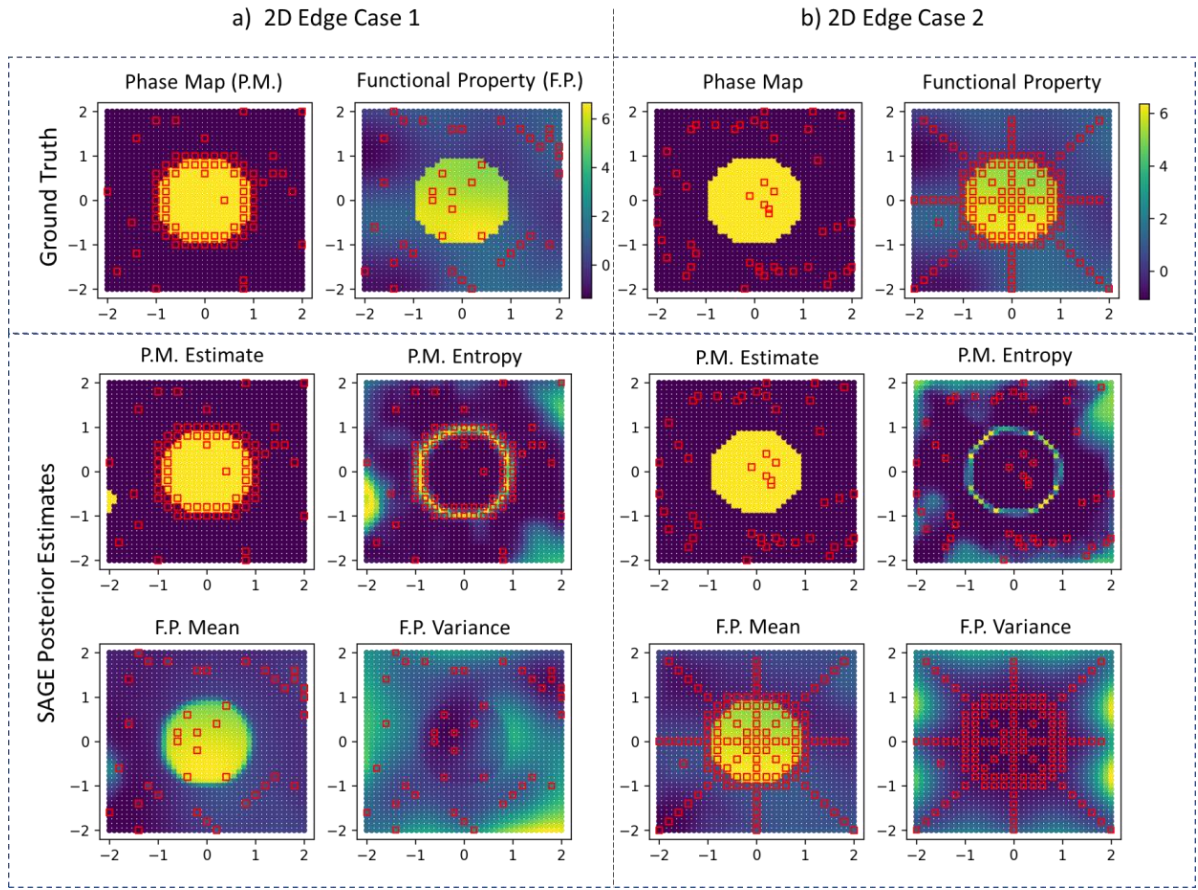

**Fig 3**. SAGE-ND demonstration on 2D example for the 2 edge cases: a) edge case 1 where structure data (red squares in phase map plots) is more informative of the phase boundary and b) edge case 2 where functional property data (red squares in functional property plots) is more informative of the phase boundary. The algorithm shows good agreement with the ground truth for both cases.

> REPLACE THIS LINE WITH YOUR MANUSCRIPT ID NUMBER (DOUBLE-CLICK HERE TO EDIT) <

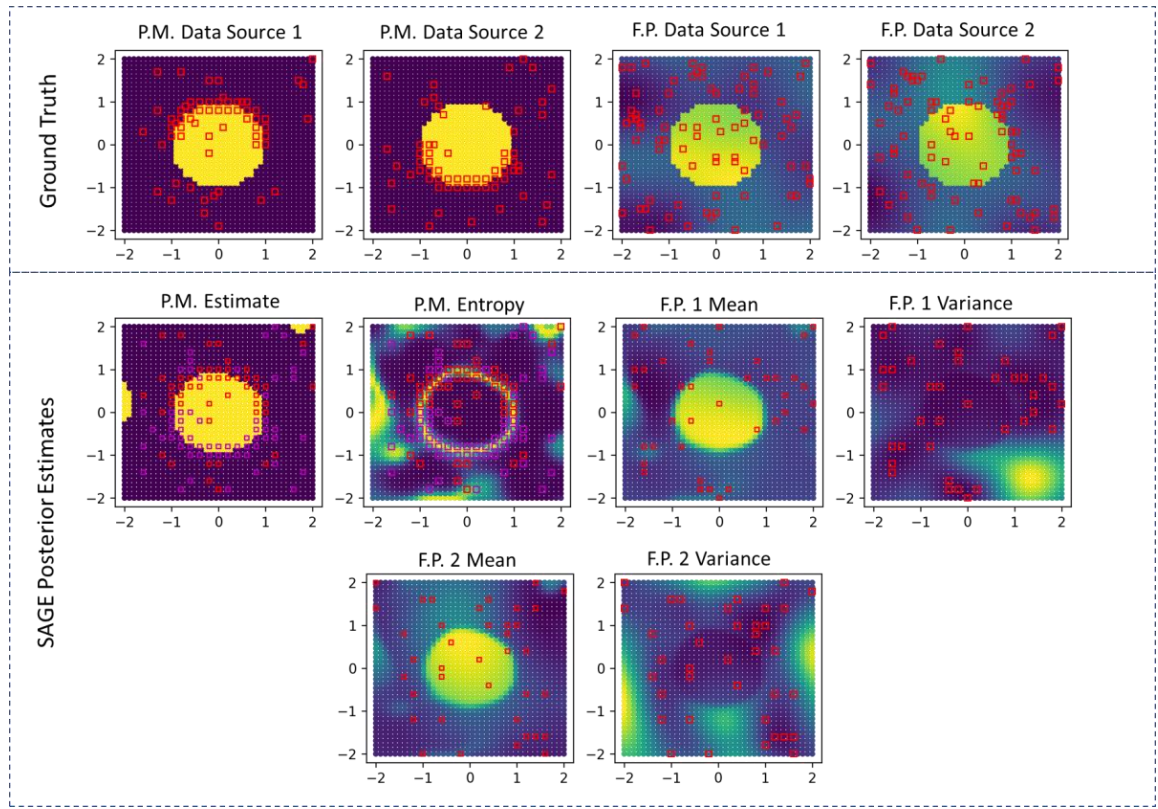

**Fig 4**. Demonstration of SAGE-ND algorithm for 2 structure data sources and 2 functional property data sources.

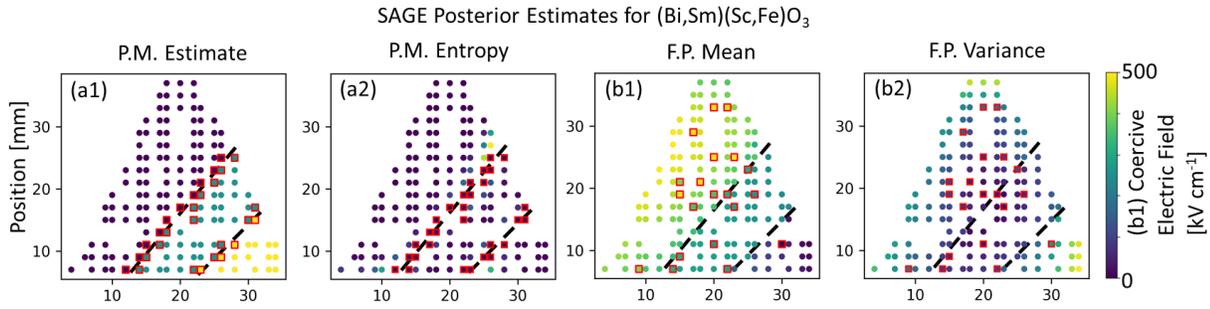

**Fig 5.** SAGE-ND applied to (Bi,Sm)(Sc,Fe)O₃ dataset. Figure 1 shows the ground truth. a1) Phase map estimate indicated by color coding with structure data indicated with red squares and phase boundaries indicated by dashed black lines. a2) entropy-measured uncertainty in the phase map of (a1), b1) CEFM estimate with functional property data indicated with red squares. b2) variance-measured uncertainty for the CEFM estimate.

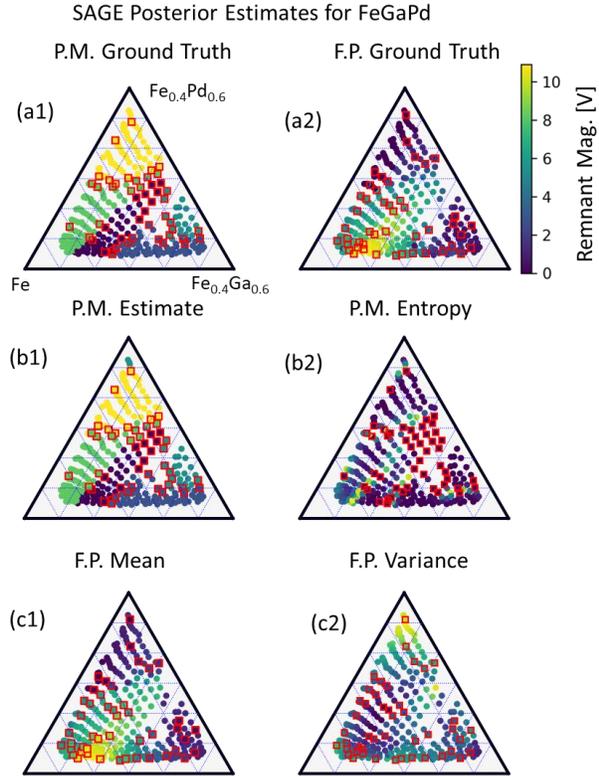

**Fig 6. SAGE-ND applied to FeGaPd dataset. a1) Phase map ground truth, a2) functional property ground truth, b1) Phase map estimate indicated by color coding, b2) entropy-measured uncertainty in the phase map of (b1), c1) remnant magnetization estimate with measured data indicated with red squares. c2) variance-measured uncertainty for the remnant magnetization estimate.**

Performance Table

| | Phase Map Performance, Accuracy [arb. units] | | | | | |
|---|---|---|---|---|---|---|
| **1D Challenges** | SAGE-1D (post. mean) | SAGE-1D-PM (post. mean) | SAGE-1D-FP (post. mean) | GP-CP (max likelihood) | GP Classification (max likelihood) | CAMEO |
| 1D Edge Case 1 | 1.00 | 1.00 | 0.97 | 0.89 | 1.00 | 1.00 |
| 1D Edge Case 2 | 1.00 | 0.90 | 1.00 | 0.97 | 0.90 | 0.92 |
| **2D Challenges** | SAGE-ND (post. mean) | SAGE-ND-PM (post. mean) | SAGE-ND-FP (post. mean) | --- | GP Classification (max likelihood) | CAMEO |
| 2D Edge Case 1 | 0.98 | 0.97 | 0.85 | --- | 0.98 | 0.86 |
| 2D Edge Case 2 | 0.98 | 0.92 | 0.97 | --- | 0.93 | 0.67 |
| (Bi,Sm)(Sc,Fe)O3 | 1.00 | 0.94 | 0.61 | | 0.89 | 0.99 |
| FeGaPd | 0.95 | 0.93 | 0.91 | | 0.99 | 0.96 |
| Functional Property Performance $R^2$ [arb. units] | | | | | | |
| **1D Challenges** | SAGE-1D (post. mean) | --- | SAGE-1D-FP (post. mean) | GP-CP (max likelihood) | --- | CAMEO |
| 1D Edge Case 1 | 0.96 | --- | 0.66 | -0.06 | --- | 1.0 |
| 1D Edge Case 2 | 1.00 | --- | 1.00 | 0.99 | --- | 0.90 |
| **2D Challenges** | SAGE-ND (post. mean) | --- | SAGE-ND-FP (post. mean) | --- | GP Regression (max likelihood) | CAMEO |
| 2D Edge Case 1 | 0.88 | --- | 0.53 | --- | 0.67 | 0.98 |
| 2D Edge Case 2 | 0.89 | --- | 0.87 | --- | 0.62 | 0.88 |
| (Bi,Sm)(Sc,Fe)O3 | 0.91 | --- | 0.27 | --- | 0.84 | 0.83 |
| FeGaPd | 0.91 | | 0.87 | | 0.90 | 0.91 |

**Table 1.** Performance scores comparing SAGE with alternative algorithms for the 1D and 2D edge cases and the real-world (Bi,Sm)(Sc,Fe)O3 and FeGaPd challenges. Here both 1 and N-dimensional Edge Case 1 has structure data that is more informative of the phase boundary and Edge Case 2 has functional property more informative of the phase boundary.